\documentclass{article}

\PassOptionsToPackage{numbers,sort&compress}{natbib}
\usepackage[preprint]{neurips_2026}

\usepackage[utf8]{inputenc} 
\usepackage[T1]{fontenc}    
\usepackage{hyperref}       
\usepackage{url}            
\usepackage{booktabs}       
\usepackage{amsfonts}       
\usepackage{nicefrac}       
\usepackage{microtype}      

\usepackage{graphicx}
\usepackage{subcaption}
\usepackage[table]{xcolor}

\usepackage{amsmath}
\usepackage{amssymb}
\usepackage{mathtools}
\usepackage{amsthm}
\usepackage{listings}

\usepackage[capitalize,noabbrev]{cleveref}

\title{Task-Induced Representational Invariances Depend on Learning Objective in Deep RL}

\author{%
  Manu Srinath Halvagal\textsuperscript{1,2,*} \quad
  Sebastian Lee\textsuperscript{1,2,3,*} \quad
  SueYeon Chung\textsuperscript{1,2,3} \\[6pt]
  \textsuperscript{1}Department of Physics, Harvard University \\
  \textsuperscript{2}Kempner Institute, Harvard University \\
  \textsuperscript{3}Center for Computational Neuroscience, Flatiron Institute \\[4pt]
  \texttt{\{manusrinathhalvagal, sebastianlee, schung\}@fas.harvard.edu} \\[4pt]
  \textsuperscript{*}Equal contribution
}

\begin{document}

\maketitle

\vspace{-1em}
\begin{abstract}
  Reinforcement Learning (RL) has long served as a model for goal-directed animal behavior in neuroscience.
  Modern deep RL has shown remarkable success across many domains,
  further strengthening this connection.
  The ability to learn abstract representations of high-dimensional state spaces underlies much of this success.
  However, theoretical understanding of these learned representations remains limited, hindering direct comparisons between models and animal learning. We address this gap by analyzing deep RL representations through the lens of MDP reduction theory.
  Investigating canonical RL algorithms in a navigation task, we find that even when performance is comparable, the value-based method (DQN) learns representations that are \emph{invariant} to MDP homomorphism symmetries, while the policy-gradient method (PPO) learns representations invariant to action symmetries.
  These differences emerge consistently across domains, have downstream consequences for transfer learning, and appear in LLMs in a prompt-dependent manner.
  Our findings provide a principled approach to comparing learned representations across RL algorithms, with demonstrated practical implications and possible insights for neural coding in the brain.
\end{abstract}
\section{Introduction}

Deep Reinforcement Learning (RL) has proved very successful in a range of tasks in recent years, including games such as Atari~\cite{mnih2015human} and Go~\cite{silver2016mastering}, and more recently in fine-tuning large language models (LLMs) to human preferences~\cite{ouyang2022training} and formal reasoning abilities~\cite{guo2025deepseek}. In neuroscience, RL has a long history as a framework for understanding animal learning~\cite{niv2009reinforcement} and is frequently used to model behaviour and explain neural data, particularly from the midbrain dopamine system~\cite{schultz1998predictive, glimcher2011understanding}. 
More generally, however, barriers remain: direct comparisons between such computational models and animal learning is 
challenging~\cite{wilson2019ten} and subject to debate across levels of analysis~\cite{krakauer2017neuroscience,gershman2024explaining}.

While there has been a recent surge of work characterizing the representations of standard deep neural networks and language models~\cite{cohen2020separability, sorscher2022neural, niu2024learning, hosseini2023large, agrawal2022alpha, ansuini2019intrinsic}, relatively little attention has been paid to understanding Deep RL algorithms in these terms. On the one hand, there is a strong literature on learning theory in tabular RL~\cite{puterman2014markov, watkins1992q, kearns2002near} that is steadily growing to incorporate function approximation of increasing generality~\cite{tsitsiklis1996analysis, jin2020provably, wang2020reinforcement}. On the other hand, the empirical community has made remarkable algorithmic and computational strides and is enjoying yet another renaissance in its post-training applications to LLMs. However, understanding model representations and how they relate to task structure has been largely overlooked, with a few notable exceptions~\cite{zahavy2016graying,wang2024investigating}.

In this work, we seek to narrow this gap by analysing RL representations through the lens of invariances to symmetries. Although some symmetries in neural representations have been documented~\cite{krupic2015grid}, 
the role of \textit{task-induced} symmetries, those arising from the structure of the behavioral problem itself, remains largely unexplored. In vision, the view that good representations progressively become invariant to nuisance variation while preserving task-relevant structure, formalized through "untangling"\cite{dicarlo2007untangling, cohen2020separability}, has been a productive bridge between computation, behavior, and neural representation. We adopt an analogous perspective for RL.
In particular, we draw on ideas from the MDP reduction literature to understand encoding of so-called homomorphic images, as well as policy symmetries in the learned representations of a number of canonical deep RL algorithms. 
We find that the representational invariances which emerge depend systematically on the learning objective: value-based methods develop invariances to MDP homomorphism symmetries, while policy-gradient methods develop invariances to action symmetries.
We perform this analysis in a range of RL domains before probing two further questions: 
do these learned symmetries have downstream functional consequences for transfer learning?
And what kinds of symmetries are present in LLM representations when solving similar simple navigation tasks framed as an interactive RL problem? Overall, our contributions are:
\begin{itemize}
    \setlength{\itemsep}{0pt}
    \item We show across a range of domains that the symmetries encoded by the representations of deep RL agents depend on the learning objective: value-based objectives (DQN) learn representations that are invariant to MDP homomorphism symmetries, while policy-gradient objectives (PPO) learn representations invariant to action symmetries, even when both solve the same underlying task.
    \item We empirically demonstrate both kinds of symmetry emerging in LLM representations when the model is prompted with different descriptions of a graph navigation task.
    \item We show empirically on selected Atari games that algorithms that learn symmetries reflecting the structure of the environment (i.e. MDP homomorphisms) transfer better than those attuned to action symmetries.
\end{itemize}

\section{Background}
\subsection{Reinforcement Learning}

The RL problem is frequently formulated as a Markov Decision Process (MDP) 
consisting of a tuple $\langle\mathcal{S},\mathcal{A},\mathcal{T},\gamma,\mathcal{R}\rangle$ where $\mathcal{S}$ is the state space, 
$\mathcal{A}$ is the action space, $\mathcal{T}$ is the state transition function, 
$\gamma$ is the discount factor and $\mathcal{R}$ is the reward function. 
RL algorithms aim to find a mapping from states to actions, 
denoted $\pi$ and known as a policy, that maximises cumulative discounted 
reward. Two canonical methods which we focus on in this work are 
DQN~\cite{mnih2013playing} and PPO~\cite{schulman2017proximal}, which sit either 
side of a common dividing line of RL algorithms; the former is a value-based method 
in which a policy is derived from a learned value-function, while the latter is a 
policy-gradient method in which the policy is directly optimised, and a 
value-function is learned as an auxiliary objective to reduce variance of the policy gradient estimates.

\subsection{MDP Reduction}

A central idea behind the present work is that of~\emph{reducing} an MDP, which can be thought of as distilling the problem to a more compact form or representation. 
MDP reduction has a long history in RL~\cite{dean1997model,ravindran2001symmetries,li2006towards,jiang2018notes} and has been
studied under various names including MDP abstraction, state abstraction and state aggregation.
The core idea is to group states or state-action pairs together based on some equivalence relation that exploits symmetries
and environment structure.
This grouping or \textit{abstraction} essentially
constructs an equivalent but smaller MDP, which is often easier to solve.

\begin{figure}[tbh]
    \centering
    \setlength{\belowcaptionskip}{-6pt}
    \includegraphics{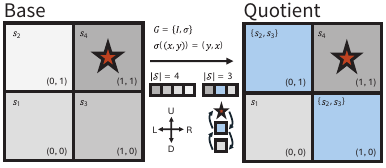}
    \caption{\textbf{MDP homomorphisms can reduce problem sizes.} \emph{(left)} toy 2x2 gridworld task, \emph{(right)} reduced MDP based on symmetry under reflection group.}
    \label{fig:reduction_schem}
\end{figure}

One might naively consider the size of an RL problem to be given by the cardinality of the state/action spaces or the complexity of the policy function class. In simple settings, these concepts overlap: when the state space is small enough and we have a discrete set of actions, we can enumerate each state-action pair and learn properties of each independently. This regime is known as tabular RL. Consider the toy example in~\autoref{fig:reduction_schem}, with 4 states and 4 actions and a single reward. Q-learning could solve this with 16 values. However, reduction techniques help uncover the core structure of the MDP, which may reveal a smaller underlying problem.

\paragraph{Bisimulation}

There are two principal groups of methods for reducing RL problems via abstraction. Under bisimulation~\cite{dean1997model}, two states $s$ and $t$ are considered bisimilar if $\forall a\in\mathcal{A}$:
\begin{align}
    r(s,a)&=r(t,a), \quad\text{and}
    \quad \sum_{s'\in C}P(s,a,s')=\sum_{s'\in C}P(t,a,s'),
\end{align}
where $P(s,a,s')$ is the probability of transitioning to state $s'$ from $s$ with action $a$, and $C$ is a set of bisimilar states. While this prescription is useful in reducing some basic environments, it is unable to account for even simple action symmetries, which are often necessary to probe functional equivalence. 
For instance, bisimulation does not reduce the 2x2 gridworld above. The fundamental reason why bisimulation fails is that it operates only over states and doesn’t allow for re-parameterisation of actions. This requires a richer theory, which we outline below.

\paragraph{MDP Homomorphism}

The method of MDP homomorphisms seeks to identify symmetries in the state-action space of the environment using ideas from group theory~\cite{ravindran2001symmetries}. For a group action applied to an MDP, the transition and reward functions are invariant, i.e. there is a symmetry under the group transformation. Formally, a group element $g\in G$ acts on the MDP via a pair of maps $(g_s, g_a)$ acting on states and actions respectively, such that for all $s,s'\in\mathcal{S}$, $a\in\mathcal{A}$, and $g\in G$:
\begin{align}
    P(s,a,s')&=P(g_s\cdot s,\, g_a\cdot a,\, g_s\cdot s'), \quad \text{and}\\ r(s,a)&=r(g_s\cdot s,\, g_a\cdot a).
\end{align}
For the 2x2 gridworld we can consider the reflection group $G=\{\sigma,I\}$ with identity $I$ and $\sigma_s((x,y))=(y,x)$. The corresponding action map is $\sigma_a(\mathrm{U})=\mathrm{R},\, \sigma_a(\mathrm{D})=\mathrm{L},\, \sigma_a(\mathrm{R})=\mathrm{U},\, \sigma_a(\mathrm{L})=\mathrm{D}$ (see ~\autoref{app:action_mapping} for verification). Under this group, states $s_2$ and $s_3$ are equivalent, as are the U/R and D/L actions respectively. This yields a reduced MDP (known as the \textit{quotient}) with abstract states $\{s_1\}$, $\{s_4\}$, $\{s_2,s_3\}$, and non-trivial abstract actions that can be described as `move towards goal' (U or R from $s_1$), `move from goal' (D or L from $s_4$), `move away from goal' (D or L from $s_2$ or $s_3$) and `move to goal' (R or U from $s_2$ or $s_3$). This reduced MDP is said to be a homomorphic image of the original MDP. Consequently, we can solve for optimality in the smaller quotient problem without loss of generality, and `lift' this solution back to the original space.
In the following, we will refer to the grouping of symmetric states based on equivalence under
an MDP homomorphism as ``MDP symmetry".

\paragraph{Policy Symmetry}
We will compare state symmetries under the reductions described above to the notion of policy or action symmetry. 
Under this framing, two states are considered symmetric if a given policy chooses the same action in both states. 
We will refer to these as ``$\pi^*$~symmetry" or ``$\pi$~symmetry" for equivalence with respect to the optimal policy or a given model policy respectively.

\subsection{Representational similarity analysis}
To compare the learned representations of different RL algorithms, we employ Representational Similarity Analysis (RSA), 
a method originally developed in neuroscience to compare neural population codes across brain regions, subjects, and species~\cite{kriegeskorte2008representational}.
RSA and extended variants such as CKA have since become standard tools for analyzing and comparing representations in both artificial neural networks and 
biological systems~\cite{kornblith2019similarity, davari2022reliability}, including for RL specifically~\cite{cross2021using}, and in regressing from brain to model representations~\cite{sucholutsky2023getting}. 
The core idea of RSA is to characterize a representation not by the individual activation patterns themselves, but by the pattern of similarities between different stimuli or states.
Specifically, for a set of $n$ states, we compute a representation similarity matrix (RSM) by measuring pairwise similarity (typically cosine similarity) between the neural activations evoked by each state. 
This yields an $n \times n$ symmetric matrix where entry $(i,j)$ reflects how similarly the network represents states $i$ and $j$.
The key advantage of RSA is that it allows us to compare representations across different network architectures, training procedures, or even different organisms, since the RSM captures the relational structure of the representation space rather than the raw activations. In our work, we use RSA to probe whether different RL algorithms encode specific symmetries in their learned representations. 
By examining the similarity structure for state pairs that are equivalent under different notions of symmetry, 
we can characterize what task structures each algorithm preferentially represents.

\subsection{Large Language Models}
LLMs have 
shown
a remarkable capability for reasoning and in-context learning~\cite{brown2020language}.
In-context learning was originally demonstrated as
a form of supervised learning without parameter updates,
but recent work has observed that LLMs also possess a capacity for in-context RL~\cite{monea2024llms}, albeit on relatively simple contextual bandit tasks.
These capabilities highlight the fact that it is possible to induce an LLM to perform goal-directed behavior purely by feeding it a description of a task.
Indeed, many interactions with a coding agent or a chat model are precisely such instances.
This emergent capability, along with the fact that RL is at the heart of LLM post-training, raises questions about what 
representations LLMs form when performing such in-context 
task execution and how these relate to representations 
learned by traditional RL.

\subsection{Other related work}

Early work on MDP reduction focused on exact abstractions in tabular settings~\cite{dean1997model, ravindran2004algebraic}, with later extensions to approximate similarity metrics~\cite{taylor2008bounding} as well as more general RL settings and constraints such as online learning~\cite{biza2018online}. 
With the advent of deep RL methods, more recent work has shifted towards continuous regimes~\cite{rezaei2022continuous, panangaden2024policy}, including attempts to directly encourage the learning of symmetry structure via equivariant architectures~\cite{van2020mdp} and sampling-based state-similarity metrics~\cite{castro2021mico}.
However, to our knowledge, ours is the first work to investigate which symmetries emerge naturally in canonical deep RL methods through representational analysis.
There have been some prior attempts to understand RL representations through empirical analysis.
Some work has analysed representations with value functions~\cite{bellemare2019geometric, dabney2021value} and policies~\cite{luss2023local} in focus, including the role of auxiliary objectives~\cite{lyle2021effect, fang2023predictive}.
Closer to our work, \citet{zahavy2016graying} looked specifically at DQN and used spatio-temporal abstractions not too dissimilar from the homomorphisms we consider to interpret representations.
However, they do not compare to other algorithms and do not ground their analysis in MDP theory. 
\citet{wang2024investigating} conducted a systematic empirical investigation measuring six representational properties (such as sparsity, orthogonality, and rank) across more than 25,000 agent-task settings to understand which properties support transfer in deep RL. While their work provides valuable insights,
they focus on measuring generic properties across auxiliary losses rather than analyzing specific task-related symmetries. 
Our work complements this by directly probing for task-induced symmetries through the lens of MDP reduction theory. 

\begin{figure*}[tbh]
    \centering
    \setlength{\belowcaptionskip}{-18pt}
    \includegraphics[width=\textwidth]{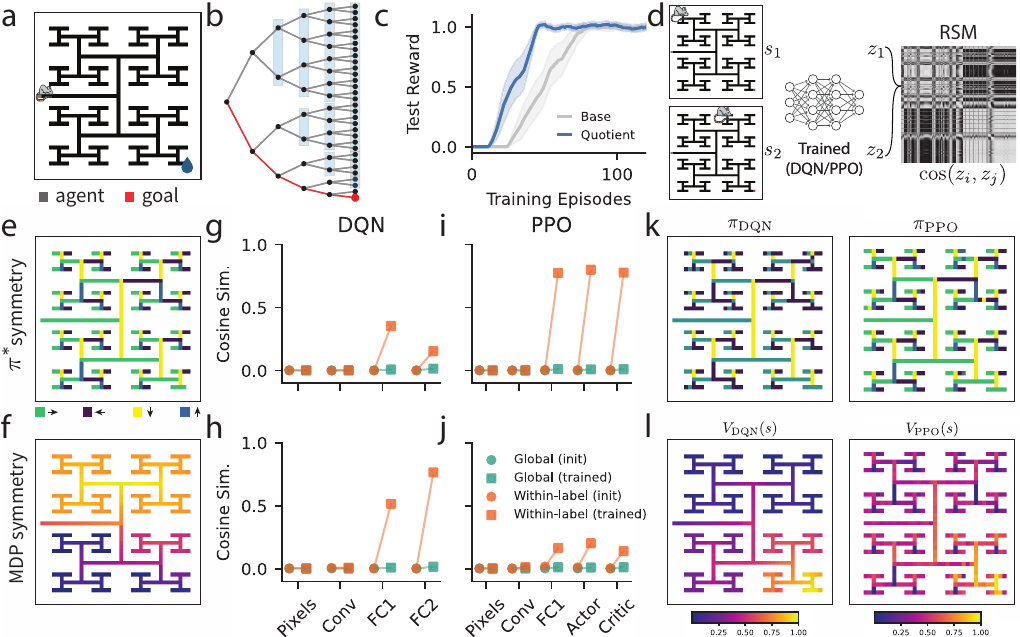}
    \caption{\textbf{Structured Navigation Task: DQN Learns MDP Homomorphisms; PPO Learns Policy Symmetry.}
    \textbf{(a)} Model environment mirroring the maze in \citet{rosenberg2021mice} consisting of six binary navigation choices to reach goal (walls white; corridors are darker paths).
    \textbf{(b)} Abstracted MDP showing binary choice points in maze as nodes in a tree graph. MDP homomorphisms induce state equivalences shown in blue.
    \textbf{(c)} Q-learning on reduced MDP (Quotient) yields faster learning compared to original MDP (Base).
    \textbf{(d)} Schematic for RSA.
    We compute cosine similarity between trained representations of all pairs of states in environment to 
    construct the full representation similarity matrix (RSM) for each model.
    Different symmetries define specific state pairings whose similarity is computed by averaging the corresponding subsets of the full RSM.
    \textbf{(e)} Visual representations of states sharing same optimal action.
    \textbf{(f)} Same as (e) but coloring indicates states that are equivalent under MDP homomorphism (cf. (a,b)).
    \textbf{(g)} Similarity structure of state representations learned by DQN.
    We compare average representational similarities between state pairs that share
    the same optimal action (within-label) before (init) and after learning (trained), with average similarities across all states (Global).
    \textbf{(h)} Same as (g) but for MDP symmetry in (f).
    \textbf{(i-j)} Same as (g-h) but for PPO.
    \textbf{(k)} Policies learned by DQN and PPO.
    \textbf{(l)} Value function learned by DQN and PPO critic.
    }
    \label{fig:meister}
\end{figure*}

\section{Representational Analysis}\label{sec:rsa}

\subsection{Symmetries in a navigation task}

A core motivation of this work was to probe
whether and how the representations learned by different
deep RL algorithms encode various task symmetries.
To that end, we first considered a labyrinth navigation task used by~\citet{rosenberg2021mice} (\autoref{fig:meister}a) to study navigation in mice.
For simulation, we formulated it as a gridworld mirroring the structure of the maze.
In addition to containing several natural abstract and spatial symmetries,
this task is well-suited for our analysis for several reasons.
Firstly, it is straightforward to represent the task both as a simple tabular environment
and an equivalent visual environment consisting of
top-down views of the labyrinth.
On the one hand, the tabular version is amenable to exact reduction with an MDP homomorphism, indicated by the blue boxes in \autoref{fig:meister}b,
making it demonstrably easier to solve (\autoref{fig:meister}c).
On the other hand, the visual environment can be used to train deep RL algorithms,
whose internal representations are unstructured at initialization but are shaped over learning.
These representations can then be probed for similarity structure based on states identified as symmetric from the tabular reduction (\autoref{fig:meister}d).
Moreover, neural data in similar navigation tasks
is routinely collected~\cite{iton2025navigraph, pedamonti2025hippocampus}, and the analysis pipeline developed here can be straightforwardly
extended to these recordings to probe analogous symmetry structures in the brain.

To investigate the symmetry structure in the learned representations, we first trained two neural networks to solve the navigation task, one with
DQN, a value-based method, and the other with PPO, a policy-gradient method, as representative canonical algorithms from each family.
The two networks had identical architectures, consisting of two convolutional layers followed by two fully connected layers. 
They differed only in the action-value head of DQN and the policy and value heads of PPO.
We trained both networks until they solved the task consistently.
Full specification of hyperparameters is provided in~\autoref{app:meister_descr}.

After training, we recorded the representations of every state in the state-space
through different layers of the two networks, centering them to zero-mean.
We then computed the cosine similarity between representations of state pairs that are symmetric under the homomorphism (MDP symmetry; \autoref{fig:meister}e) and the optimal policy ($\pi^*$~symmetry; \autoref{fig:meister}f). 
We observed significant differences in the similarity structure
learned by the two algorithms.
Specifically, DQN yielded low similarity between pairs of states sharing the same optimal action, but high similarity between pairs of states
that are MDP symmetric (\autoref{fig:meister}g,h).
In contrast, PPO yielded high similarity for the policy pairs and low similarity for the MDP symmetry (\autoref{fig:meister}i,j).
Lastly, we verified that the particular similarity structures are present in states across most of the maze hierarchy (\autoref{sec:meister_suppl}).

These differences are striking since both models have identical network architectures up to the readout, receive the same inputs and
both successfully solve the navigation task (as shown by optimal action readouts in \autoref{fig:meister}k).
However, the results are also consistent with the
underlying learning objectives of the two methods.
DQN is a value-based RL method and therefore learns a good encoding of the value function, which for this particular environment induces a symmetry structure that coincides with the MDP homomorphism.
On the other hand, PPO is predominantly a policy optimization approach and therefore tends to prioritize organizing its representations according to the optimal action over other equivalences. This finding is further verified by the higher fidelity of the value function learned by DQN vs. the PPO critic in \autoref{fig:meister}l.
Notably, in PPO the encoder is shared between the actor and critic heads; the value signal from the critic is evidently dominated by the policy-gradient signal from the actor, resulting in a representation organized around action equivalences rather than value equivalences.

\paragraph{Partition overlap.} A natural concern is that the dissociation between algorithms could be artifactual if the MDP-symmetric and policy-symmetric partitions overlap heavily. We therefore quantified the partition geometry directly: only 2.4\% of state pairs are MDP-symmetric, vs. 30\% policy-symmetric (i.e., share optimal action). Crucially, 62\% of MDP-equivalent pairs also share the optimal action, but only 5\% of policy-equivalent pairs are MDP-equivalent. This asymmetry has two consequences. First, DQN will inevitably exhibit a residual policy-similarity signal (visible in \autoref{fig:meister}g) simply because most MDP-equivalent pairs happen to share actions---a property of the gridworld, not the algorithm. Second, the converse does not hold: PPO's elevated policy similarity cannot be explained by MDP overlap, since policy-equivalent pairs are almost never MDP-equivalent. The dissociation between value-based and policy-gradient representations is therefore not an artifact of confounded partitions.

\paragraph{Isolating the learning objective with Deep SARSA($n$)}

A key question is whether the representational differences above arise from the 
learning objective (value-based vs.\ policy-gradient) or from other algorithmic 
differences, such as data collection strategy (off-policy replay for 
DQN vs.\ on-policy rollouts for PPO). To directly test this, we trained Deep 
SARSA($n$)~\cite{rummery1994line}, which is \emph{on-policy} like PPO but 
\emph{value-based} like DQN. We trained SARSA($n$) on a depth-4 version of the 
labyrinth with random start positions and potential-based reward shaping to 
facilitate convergence; DQN and PPO were retrained on the same depth-4 maze for a 
fair comparison. Full details are in~\autoref{app:sarsa_descr}. The results confirm our prediction: SARSA($n$) encodes MDP homomorphism symmetry 
like DQN, not policy symmetry like PPO (\autoref{fig:sarsa_results}), despite 
sharing PPO's on-policy data collection. This strengthens the evidence that the 
\emph{learning objective}---value-based vs.\ policy-gradient---is the primary driver 
of representational symmetry structure, rather than the mode of data collection.
\paragraph{Symmetry emergence over training}
We also tracked how symmetry structure develops during training for both 
DQN and PPO (\autoref{sec:symmetry_dynamics}). Both 
algorithms show a broad initial increase in pairwise similarity across all 
state pairs, followed by selective retention consistent with each algorithm's 
learning objective. Crucially, the characteristic symmetry pattern for each algorithm emerges \emph{before} task performance peaks, suggesting that 
representational organization precedes, rather than follows, behavioral 
optimization.

\subsection{Bilateral symmetries in more complex environments}

To test whether the dissociation generalizes beyond exactly-reducible environments, we performed analogous analyses in Cartpole (continuous, low-dimensional) and Pong (high-dimensional pixel observations). Both admit a heuristic bilateral symmetry---left-right for Cartpole and top-bottom for Pong---but lack tractable exact MDP homomorphic reductions; we therefore used mirrored state pairs as an incomplete proxy for MDP symmetry, and defined policy symmetry operationally as states for which the learned policy selects the same action. The qualitative pattern from the gridworld replicates in both environments: only DQN representations show elevated similarity for mirrored state pairs relative to baseline, while PPO representations encode same-action similarity most strongly in the later layers (actor head in Pong). The MDP-symmetry signal is weaker than in the labyrinth, particularly for Pong, which may partly reflect the 
local receptive fields of convolutional networks; architectures with explicit global structure (e.g.\ equivariant convolutions or spatial attention) might encode global bilateral symmetries more efficiently. 
Moreover, bilateral symmetries represent part of a potentially much larger set of symmetries in the underlying MDP, limiting the insights one could draw from a simple representation analysis.
To go further, we performed targeted transfer learning experiments detailed in the following.
Detailed methodology and full results are in~\autoref{app:bilateral_symmetries}.

\section{Transfer Learning}

\begin{figure}[tbh]
    \centering
    \setlength{\belowcaptionskip}{-6pt}
    \includegraphics[width=0.85\textwidth]{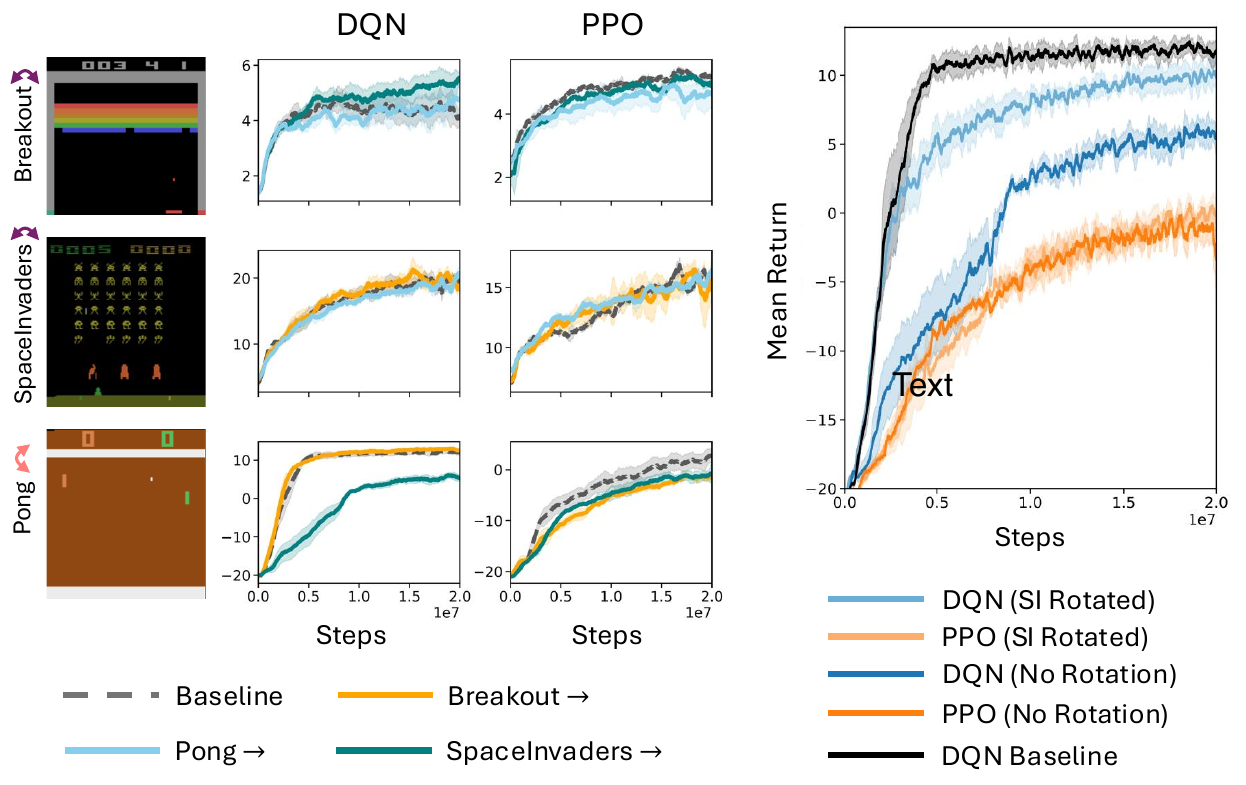}
    \caption{\textbf{Representing MDP Homomorphisms Can Aid Transfer Learning in Atari.} \emph{(left)} Sample screens of the three games: Breakout, SpaceInvaders, Pong (top-bottom). All three games have global notions of reflective symmetry. \emph{(middle)} episode returns over the course of training for DQN and PPO on the three games. Grey-dashed curves are baseline runs trained on a single task. Coloured curves for a given model/game combination are pre-trained on the game indicated by the legend. Overall DQN, which learns MDP homomorphisms, outperforms PPO, which learns policy symmetry. Results are averaged over 3 seeds. Full details can be found in~\autoref{app:transfer_descr}. (\emph{right}) Mean return comparison between agents trained with standard observations and agents trained with SI-rotated observations. DQN, which encodes MDP homomorphisms, maintains performance under observation rotation while PPO, which encodes policy symmetries, shows greater sensitivity.
    }
    \label{fig:transfer_rebuttal_combined}
\end{figure}

In all experiments in~\autoref{sec:rsa}, every model we analyze solves the given task optimally despite significantly different representations. However,
different learned symmetry structures can change generalisation and performance downstream. We demonstrate this on a suite of transfer learning scenarios in Atari. 
Learning symmetry structures such as reflection or translational invariance in an environment can translate from task to task; this has been proposed as a model of learning and memory in animals~\cite{whittington2020tolman}.
Conversely, action equivalence requires more fine-grained remapping across tasks. 
Accordingly, we hypothesised that DQN would perform better than PPO in transfer between Atari tasks that share reflection symmetry. Results for the experiment are shown in~\autoref{fig:transfer_rebuttal_combined}. The left hand side shows the three Atari games we select: Breakout (Vertical), SpaceInvaders (Vertical) and Pong (Horizontal); all of which contain strong reflective symmetry in their environments. We take robust stable-baseline implementations of DQN, PPO and QRDQN~\cite{stable-baselines} and train each sequentially on every pairwise combination of games. Since the games do not have the same action spaces, we re-initialise the head layers with possibly different dimensions at task boundaries. We then compare performance of the baseline model (i.e.. trained only on a given game) with the transfer performance where the model was previously trained on a different game. It is well documented that deep RL algorithms suffer from catastrophic loss of plasticity~\cite{abbas2023loss}, so we focus here on early parts of training where we expect the relevant symmetry structures to be learned without models moving into the over-fitting or non-plastic regime. Further results on the full course of training can be found in~\autoref{fig:fulltransfer}. 

We checkpoint our baseline runs at 10 million steps and load these weights into the transfer task. \autoref{fig:transfer_rebuttal_combined} and further data in~\autoref{tab:stats} in~\autoref{app:transfer_descr} show that over the first 20 million steps of training there is more improvement for DQN (and QRDQN), while PPO mostly sees no or negative transfer. Interestingly the one clear failure mode for DQN and QRDQN is when transferring from SpaceInvaders to Pong, where the symmetry flips from vertical to horizontal while strongest transfer is from SpaceInvaders to Breakout where the symmetry direction is upheld. Full experimental details including hyper-parameters can be found in~\autoref{app:transfer_descr}. We note that these results demonstrate a correlation between MDP-homomorphism encoding and transfer performance; a definitive causal test (e.g.\ training PPO with an auxiliary MDP-symmetric loss to induce homomorphism encoding) remains future work.

To further illustrate the downstream consequence of MDP-homomorphism encoding, \autoref{fig:transfer_rebuttal_combined} compares performance of agents trained on observations with and without symmetry-invariant (SI) rotations applied. Agents whose representations encode MDP homomorphisms (DQN) show more stable performance under such observation transformations, while policy-gradient agents (PPO) are more sensitive, consistent with their weaker structural encoding of the environment.


\section{In-context Task Execution with an LLM}

\begin{figure}[tbh]
    \centering
    \setlength{\belowcaptionskip}{-8pt}
    \includegraphics{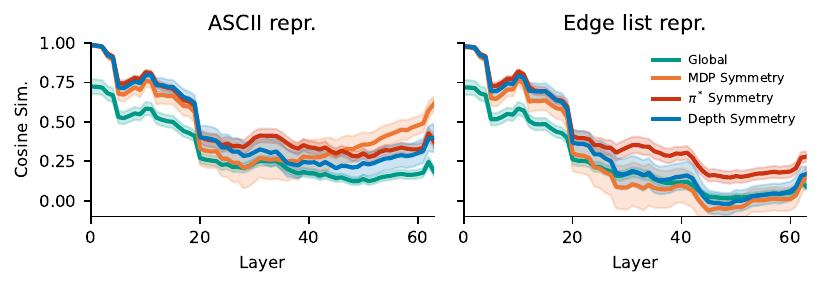}
    \vspace{-20pt}
    \caption{\textbf{Representational symmetry structure in an LLM solving the abstracted MDP of the labyrinth (cf.\ Fig.~\ref{fig:meister}b).}
    Cosine similarity between all pairs of states (Global), states symmetric under an MDP homomorphism (Within MDP), states sharing the same optimal action (Within policy),
    and states at the same depth in the MDP tree (Within depth).
    The LLM representations show high similarity for state pairs under policy symmetry for any description of the task.
    High similarity for the MDP symmetry above that for the naive depth symmetry only emerges
    when the underlying graph is described using an ASCII representation (top),
    but not when described with a list of edges (bottom).
}
    \label{fig:llm_sim}
\end{figure}

Having established that different RL algorithms encode fundamentally different symmetries despite solving tasks optimally, we investigated whether similar representational differences emerge in LLMs performing in-context task execution. We prompted the Qwen2.5-72B-Instruct model~\cite{qwen2,qwen2.5} to solve the labyrinth navigation task, varying the format in which the abstract MDP was described: an ASCII tree (making the hierarchical structure visually explicit) or an edge list (same information, less structured). Full experimental details are in~\autoref{app:llm_descr}. The results (\autoref{fig:llm_sim}) reveal a striking prompt-dependence: policy-level symmetry is robustly encoded regardless of format, mirroring what we observed for PPO. MDP homomorphism symmetry above the naive depth baseline only emerges with the ASCII format, suggesting that structural invariances can be extracted when the prompt makes them explicit. A broader survey of six prompt formats is shown in \autoref{fig:llm_sim_full}. These results suggest that LLMs can represent both symmetry types simultaneously, with the prominence of each modulated by the input format---a flexibility not present in trained RL algorithms. We discuss implications for LLM post-training and RLHF in~\autoref{app:llm_discussion}.

\section{Discussion}

In this work, we have investigated the representations learned by canonical 
deep RL algorithms through the lens of MDP reduction theory. 
We find that value-based methods such as DQN preferentially encode 
MDP homomorphism symmetries, while policy-gradient methods such as 
PPO encode policy symmetries. These differences emerge consistently 
across a range of domains including navigation, continuous control and 
Atari. Downstream, these representational signatures are likely to have significant consequences; we demonstrate this in one regime by showing that transfer learning is enhanced for models that tend to learn MDP homomorphisms compared to those that learn policy symmetries.
Furthermore, we find that LLMs exhibit both kinds of symmetries in a 
prompt-dependent manner when solving similar navigation tasks.

These findings open several avenues for future work.
From a machine learning perspective,
it would be valuable to explore how to
encourage specific symmetry structures in the representations regardless of the learning algorithm.
From an engineering standpoint, our results suggest that the choice of RL algorithm has predictable consequences for the type of structure encoded: practitioners seeking representations that capture environment symmetries (e.g.\ for transfer or generalization) may prefer value-based methods, while those prioritizing action-level organization may prefer policy-gradient methods.
Model-based RL approaches that use auxiliary losses to learn the environment structure directly, but not
necessarily compactly with respect to symmetries, would be an interesting place to begin.
Building on our experiments in transfer learning, a more thorough understanding of the impacts of these representational differences on various learning regimes such as fine-tuning, continual learning and other kinds of generalisation could help us build more robust and flexible models.
A natural further step would be to probe whether the representational invariances we document are truly equivariant---i.e., whether there exists a linear map $W$ such that $z(g_s\cdot s) \approx W\cdot z(s)$---by fitting such a regression and examining $R^2$ and the structure of $W$ across algorithms. We leave this as a concrete direction for future work.

From a neuroscience perspective,
it would be interesting to investigate whether similar symmetry structures
are present in neural data from animals performing comparable tasks,
and what this reveals about learning and computation in the brain.
The analysis pipeline developed here
can be directly extended to neural recordings.
Symmetry is already known to play a fundamental role in neural spatial representations~\cite{krupic2015grid}. 
Our framework provides a natural way to probe whether biological neural circuits preferentially encode MDP-like task symmetries versus policy-like action mappings. 
Testing such predictions would require simultaneous recordings from multiple regions during tasks with known symmetry structure, precisely the kind of labyrinth navigation paradigm we analyzed here. 
The recent development of high-density neural recordings in navigating animals~\cite{steinmetz2021neuropixels} makes such investigations increasingly feasible.

RL has recently become central to aligning LLMs with 
human preferences~\cite{ouyang2022training} and improving their reasoning capabilities~\cite{guo2025deepseek}.
In Reinforcement Learning from Human Feedback (RLHF), models are 
first fine-tuned on human preference data to learn a reward model, 
which is then used to train the policy via RL algorithms~\cite{ouyang2022training}. 
These methods have proved crucial for training models with 
enhanced reasoning abilities, as demonstrated in systems like 
DeepSeek-R1~\cite{guo2025deepseek}, where RL is used to encourage 
step-by-step reasoning and verification behaviors. 
However, it is unknown whether and how the RL
post-training changes the LLM representations, potentially in ways detrimental to downstream in-context learning similar to our observations for PPO transfer on Atari tasks.

Our scope of empirically evaluating the signatures of different symmetries echoes  
early work characterizing representational structures via similarity~\cite{kriegeskorte2008representational}.
In vision, this inspired analytical theories of neural manifold geometry underlying invariant object recognition \cite{chung2018classification, chou2024geometry}, which have revealed consequences for e.g. representational efficiency \cite{yerxa36efficient}. 
We hope our findings here could similarly serve as a starting point for theories of RL representation geometry.

\textbf{Limitations.}
Our analysis has several constraints.
First, our RSA approach assumes a one-to-one mapping from observations to states, which breaks down under partial observability (POMDPs), recurrent architectures, or model-based methods that maintain explicit world models---these settings would require extensions of our framework.
Second, while our Deep SARSA($n$) experiment isolates the learning objective from the data collection strategy, further algorithmic axes remain unexplored (e.g.\ SAC, successor representations, model-based RL).
Third, our analysis uses only cosine similarity; other metrics (CKA, mutual information) might reveal complementary structure.
Furthermore, our large-scale experiments on Atari transfer and LLMs are limited in the extent of the models and tasks studied, but are good starting points for more thorough investigations. Similarly, broader environmentation with continuous morphological symmetries  (beyond CartPole in~\autoref{app:cartpole_descr}) such as HalfCheetah or Ant in MuJoCo represent a natural extension beyond the predominantly discrete symmetries studied here.
Finally, while we document empirical representational differences, we lack a formal theory to predict which symmetries emerge from a given algorithmic property.
These limitations suggest that our findings should be viewed as an initial empirical characterization rather than a complete theory.

In summary, we have provided a principled framework grounded in MDP 
reduction theory for comparing learned representations across RL algorithms.
This framework has direct implications for algorithm selection in transfer 
learning and continual RL, for interpreting and improving LLM post-training, 
and for understanding task representations in neuroscience. 

\section*{Acknowledgments}

We thank Will Slatton and Chanwoo Chun for helpful comments and all members of the Chung lab for early feedback.
This work was supported by the NIH R01 grant DA059220 (to S.C. and M.S.H.), as well as by a Sloan Research Fellowship and a Klingenstein-Simons Award (to S.C.). The Flatiron Institute is a division of the Simons Foundation.

\bibliography{main}
\bibliographystyle{plainnat}


\clearpage

\appendix

\section{Navigation Symmetries along Maze Depth}
\label{sec:meister_suppl}

Below we document some further intriguing results around the strength of similarity as a function of depth in the abstracted MDP tree corresponding to the labyrinth navigation task (cf.~\autoref{fig:meister}b). Both PPO and DQN show distinct trends along depth, which relates to ongoing research agendas in compositional and hierarchical learning~\cite{botvinick2012hierarchical, lake2017building, saxe2019mathematical, cohen2020separability}, and may be worthy of future study.

\begin{figure*}[tbh]
    \centering
    \includegraphics{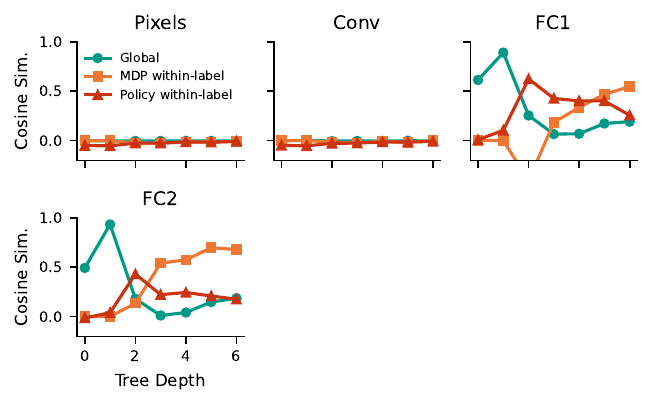}
    \caption{\textbf{DQN representations encode MDP symmetry across maze depth.} 
Cosine similarity between state pairs grouped by MDP symmetry and 
optimal policy symmetry for DQN, shown separately for states at each depth level 
in the navigation tree (cf.~\autoref{fig:meister}b). Global similarity shown for comparison. MDP symmetry 
is elevated above global similarity in later network layers (FC1, FC2) and increases with tree depth, while policy symmetry remains near baseline.
States at deeper tree depths are more numerous and therefore dominate the aggregate measure plotted in \autoref{fig:meister}g,h.}
    \label{fig:meister_depthwise_dqn}
\end{figure*}

\begin{figure*}[!h]
    \centering
    \includegraphics{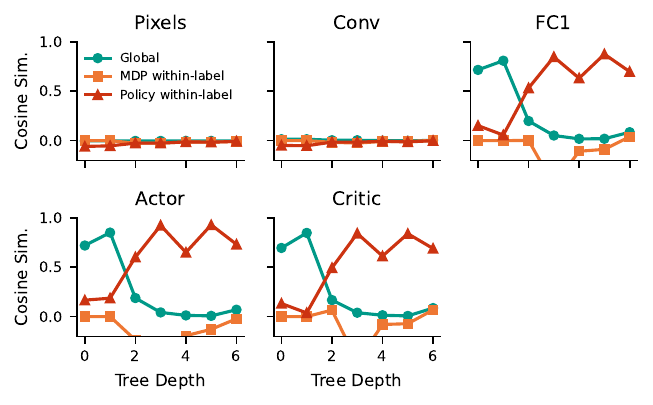}
    \caption{\textbf{PPO representations encode policy symmetry across maze depth.}
    Same as \autoref{fig:meister_depthwise_dqn} but for PPO. In contrast to DQN, PPO shows 
elevated similarity for states sharing the same optimal action across tree 
depths, while MDP symmetry remains near global baseline levels. This pattern 
is consistent across the shared FC layers and persists into 
both the Actor and Critic heads, indicating that policy-based organization dominates 
PPO representations regardless of state location in 
the task structure.
    }
    \label{fig:meister_depthwise_ppo}
\end{figure*}

\section{Action-Space Mapping for the 2$\times$2 Gridworld}
\label{app:action_mapping}

For the 2$\times$2 gridworld with reflection group $G=\{\sigma, I\}$ and $\sigma_s((x,y))=(y,x)$, the corresponding action map $\sigma_a$ is defined as:
\[
\sigma_a(\mathrm{U}) = \mathrm{R}, \quad \sigma_a(\mathrm{D}) = \mathrm{L}, \quad \sigma_a(\mathrm{R}) = \mathrm{U}, \quad \sigma_a(\mathrm{L}) = \mathrm{D}.
\]
We verify that Equations~3--4 hold under this group action for all state-action pairs. Consider the transition from $s_2=(0,1)$ taking action $\mathrm{U}$, which leads to $s_4=(1,1)$. Under $\sigma$, we have $\sigma_s(s_2)=s_3=(1,0)$ and $\sigma_a(\mathrm{U})=\mathrm{R}$. Taking action $\mathrm{R}$ from $s_3=(1,0)$ also leads to $\sigma_s(s_4)=(1,1)=s_4$, confirming $P(s_2,\mathrm{U},s_4)=P(s_3,\mathrm{R},s_4)$. Reward invariance follows similarly since reward is only given at the goal state $s_4$, which is fixed under $\sigma_s$.

\section{Deep SARSA($n$) Experiment}
\label{app:sarsa_descr}

\subsection{Setup}

Deep SARSA($n$) is on-policy like PPO but value-based like DQN, making it a critical control condition for isolating the role of the learning objective from the data collection strategy. We trained SARSA($n$) on a depth-4 labyrinth (smaller than the depth-6 maze in the main experiments) with random start positions, which are necessary for the RSA analysis to sample representations across the full state space. We applied potential-based reward shaping~\cite{ng1999policy} using the BFS distance to the goal as the shaping potential, to facilitate reward propagation under SARSA's single-step updates. DQN and PPO were retrained on the same depth-4 maze with random starts for a fair comparison. Network architecture and other hyperparameters followed~\autoref{app:meister_descr}.

\subsection{Results}

\begin{figure*}[tbh]
    \centering
    \begin{minipage}[c]{0.30\textwidth}
        \centering
        \includegraphics[width=\textwidth]{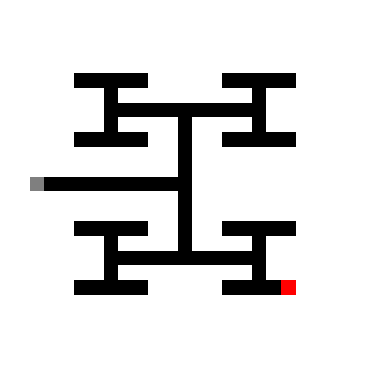}
    \end{minipage}
    \hfill
    \begin{minipage}[c]{0.68\textwidth}
        \centering
        \includegraphics[width=\textwidth]{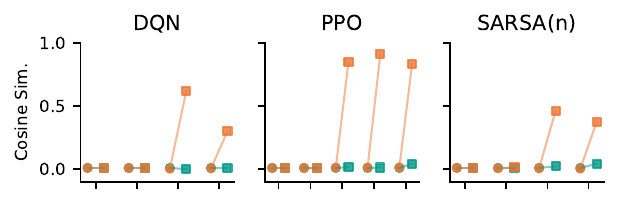}
        \vspace{4pt}
        \includegraphics[width=\textwidth]{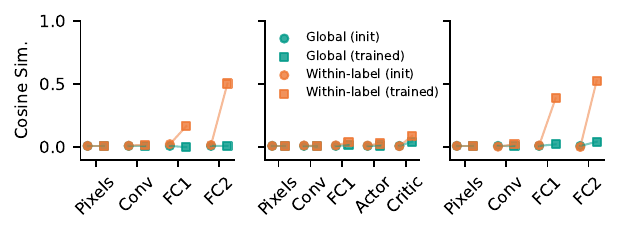}
    \end{minipage}
    \caption{\textbf{Deep SARSA($n$) encodes MDP homomorphism symmetry like DQN, not policy symmetry like PPO.} \emph{Left:} Depth-4 labyrinth environment used for the three-way comparison. \emph{Right, top:} Policy symmetry. PPO shows elevated within-class similarity after training; DQN and SARSA($n$) show moderate elevation. \emph{Right, bottom:} MDP homomorphism symmetry. DQN and SARSA($n$) show elevated within-class similarity (orange, squares) above global baseline (teal) after training; PPO does not. Circles = initialization, squares = after training. SARSA($n$) is on-policy (like PPO) but value-based (like DQN), isolating the learning objective as the key driver of representational symmetry.}
    \label{fig:sarsa_results}
\end{figure*}

As shown in~\autoref{fig:sarsa_results}, SARSA($n$) encodes MDP homomorphism symmetry in the same manner as DQN, and shows only weak policy symmetry---consistent with DQN and in contrast to PPO. This confirms that the value-based learning objective, rather than off-policy data collection, is the primary driver of MDP homomorphism encoding in neural representations.

\clearpage
\section{Symmetry Dynamics over Training}
\label{sec:symmetry_dynamics}

\begin{figure*}[tbh]
    \centering
    \begin{subfigure}[t]{0.48\textwidth}
        \centering
        \includegraphics[width=\textwidth]{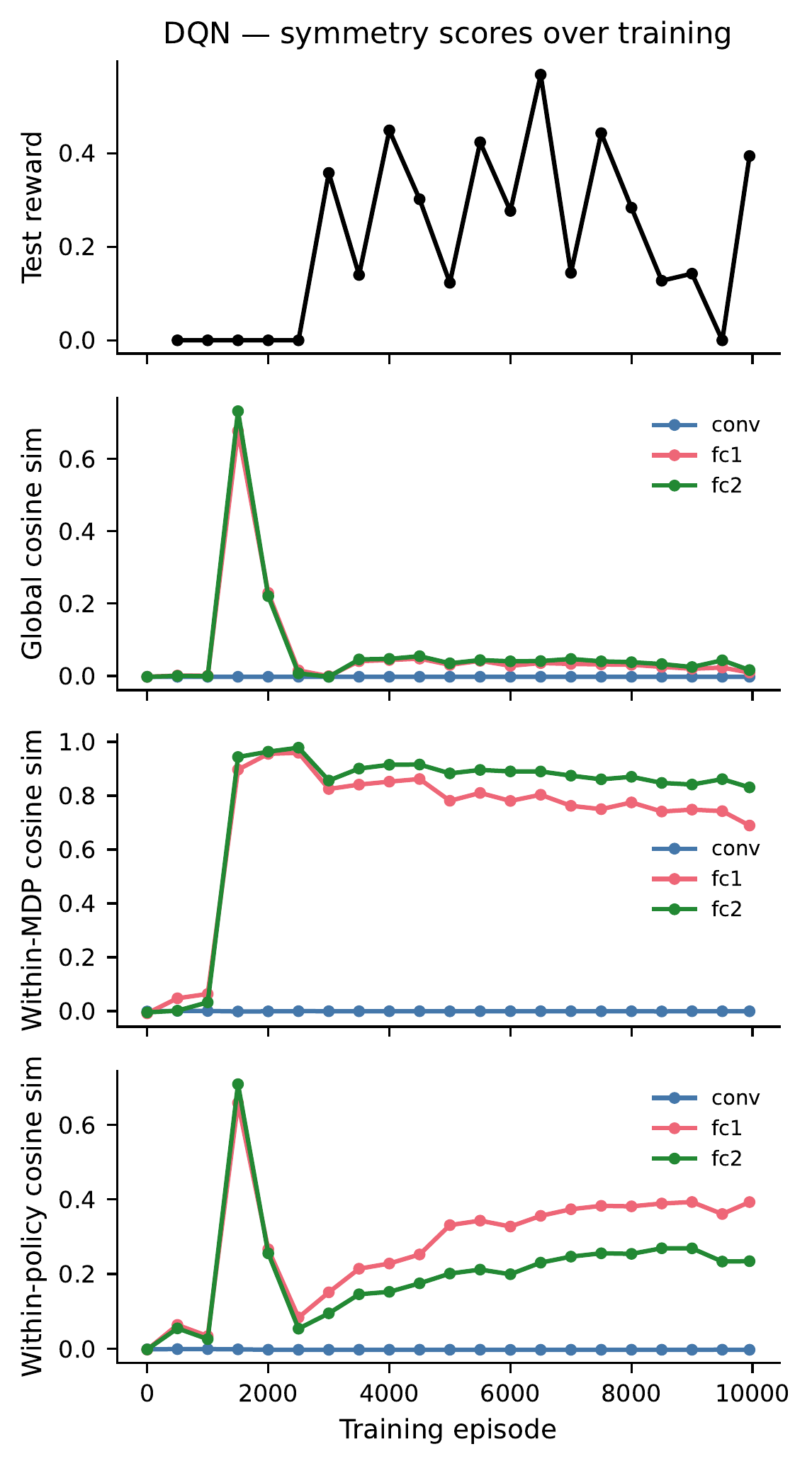}
        \caption{DQN: MDP symmetry increases over training while policy symmetry remains near baseline.}
    \end{subfigure}
    \hfill
    \begin{subfigure}[t]{0.48\textwidth}
        \centering
        \includegraphics[width=\textwidth]{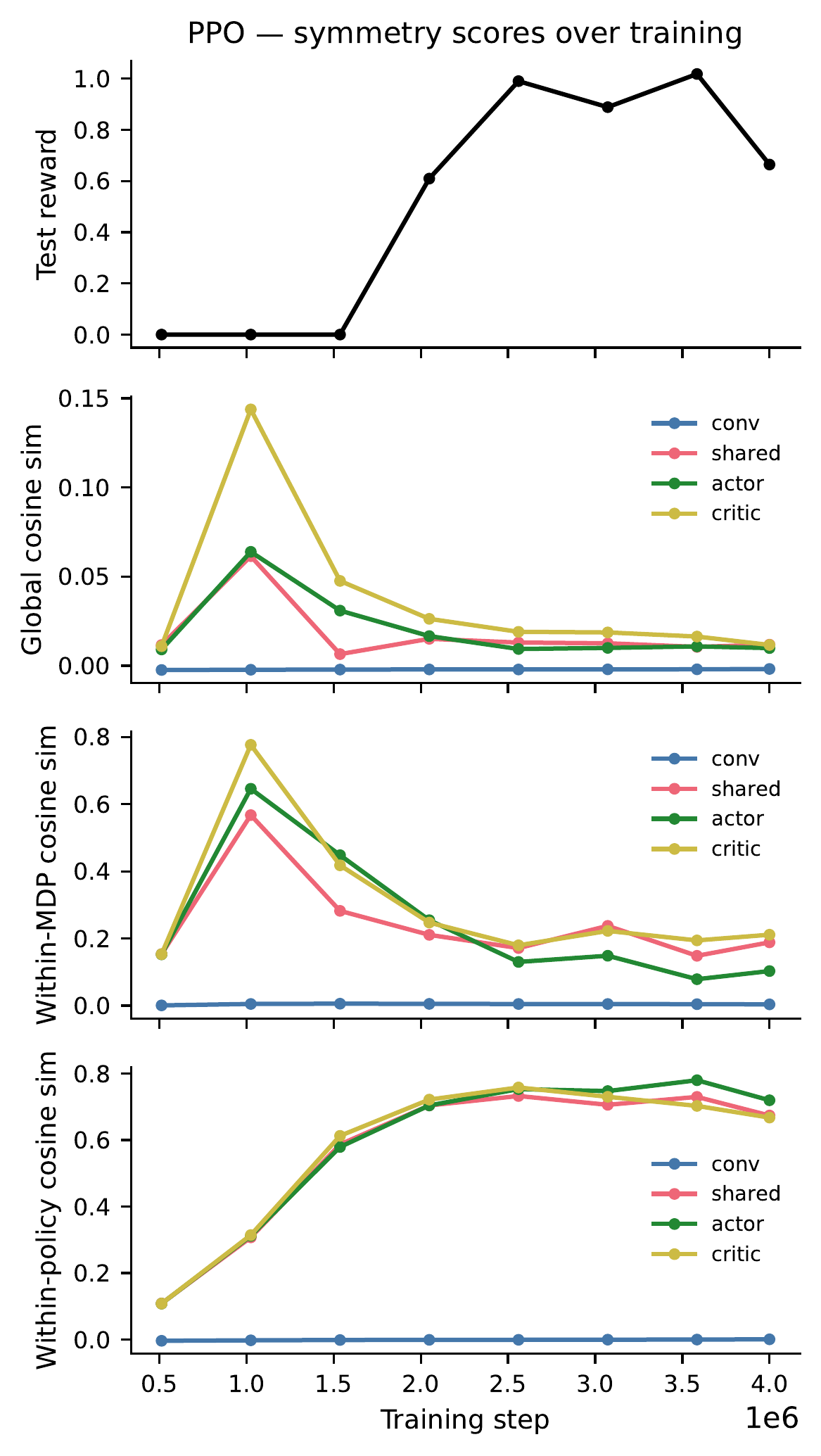}
        \caption{PPO: Policy symmetry increases over training while MDP symmetry remains near baseline.}
    \end{subfigure}
    \caption{\textbf{Symmetry structure emerges during training, prior to performance peaking.} Cosine similarity within MDP-symmetric pairs (orange) and policy-symmetric pairs (blue) tracked across training for DQN (left) and PPO (right). Both algorithms show an initial broad increase in similarity, followed by selective retention. The characteristic symmetry pattern for each algorithm emerges before task performance peaks.}
    \label{fig:symmetry_dynamics}
\end{figure*}

\section{Bilateral Symmetries in CartPole and Pong}\label{app:bilateral_symmetries}

We wanted to investigate whether the differences in symmetry encoding observed in the gridworld generalised to more complex environments. To that end, we undertook a similar analysis for Cartpole, a simple low-dimensional environment but with a continuous state-space, and Pong, a very high-dimensional pixel-based environment. These environments are difficult to reduce exactly via an MDP homomorphism, yet they both share a bilateral symmetry: left-right for Cartpole and top-bottom for Pong. Due to the lack of an exact reduction, we investigated the similarity structure by limiting ourselves to these manually identified heuristic state symmetries. Furthermore, unlike in the labyrinth, ground-truth optimal policies are not available in these environments. We therefore defined policy symmetry operationally as states for which the learned policy $\pi$ selects the same action.

\begin{figure*}[tbh]
    \centering
    \setlength{\belowcaptionskip}{-6pt}
    \includegraphics[width=\textwidth]{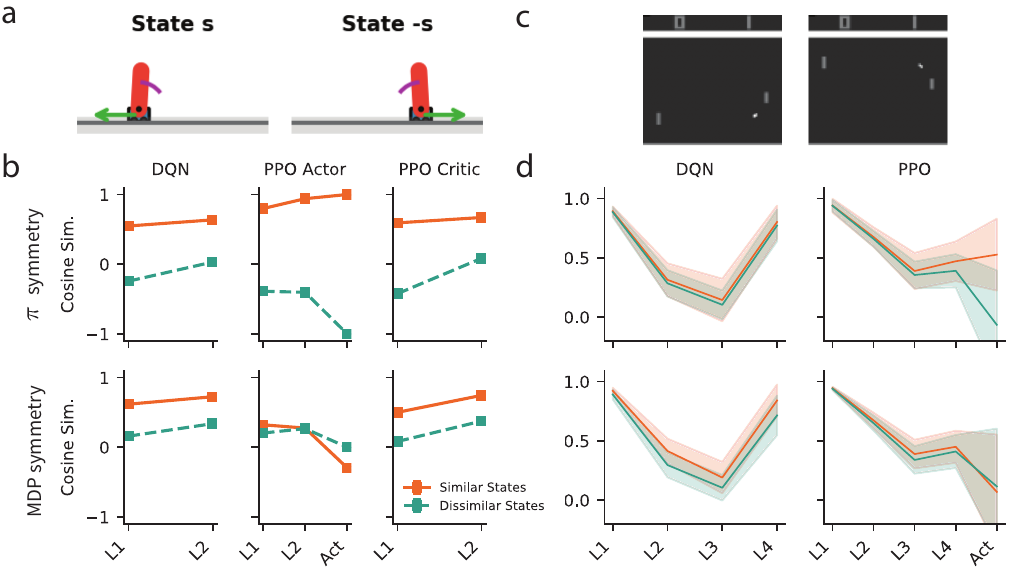}
    \caption{\textbf{Algorithm-Dependent Symmetry Representations Persist Across RL Domains.}
    \textbf{(a)} A pair of symmetric states in the Cartpole environment, defined as mirror images of each other about the center.
    Such pairs are a subset of possible equivalent states under
    an MDP homomorphism.
    \textbf{(b)} Learned similarity structure with respect to policy symmetry (top) and the mirror symmetry (bottom) across network layers after learning with DQN and PPO.
    Both DQN and PPO show high representation similarity for pairs
    of states that share the same action under the learned policy,
    and negative similarity for pairs that do not,
    indicating a possible linear encoding of the policy in both cases.
    Only DQN shows higher similarity for mirrored pairs of states,
    compared to random pairs of states in the environment.
    \textbf{(c)} Same as (a) but for Pong.
    \textbf{(d)} Same as (b) but for networks trained on Pong. The Pong panel shows representations from the shared convolutional encoder (actor and critic share the backbone, differing only at the final heads).
    DQN representations do not show a higher similarity for states sharing the same action under the learned policy, but do show slightly higher similarity for mirrored pairs of states when compared to baseline similarity between random pairs of states.
    PPO representations show higher similarity for states sharing the same action, but only close to the end of the network hierarchy with the most separation at the actor.
    PPO representations do not have higher similarity for mirrored state pairs.
    }
    \label{fig:left_right}
\end{figure*}

For Cartpole, we again trained DQN and PPO to solve the task. We then sampled 500 random states and their mirrors (\autoref{fig:left_right}a) and computed the cosine similarity between pairs of mirrored states as a proxy for the true MDP symmetry. We defined $\pi$ symmetry by grouping states that shared the same action under the learned policy. The analysis revealed that both DQN and PPO learned representations with high similarity for state pairs sharing the same action and negative similarity for pairs requiring the opposite action (\autoref{fig:left_right}b). This suggests a possible linear encoding of the policy in both networks for this environment. However, only DQN showed elevated similarity for mirrored state pairs compared to random baseline pairs, consistent with value-based methods capturing underlying MDP structure.

For Pong, we performed a similar analysis using the top-bottom bilateral symmetry inherent to the game (\autoref{fig:left_right}c). This environment presents a considerably more challenging test case due to its high-dimensional pixel observations and more complex dynamics. We analysed networks trained with DQN and PPO from the Stable Baselines3 model zoo, and then sampled states along with their vertically mirrored counterparts. The analysis (\autoref{fig:left_right}d) revealed a similar pattern as the other environments. DQN representations did not show elevated similarity for states sharing the same action under the learned policy, but did exhibit slightly higher similarity for mirrored state pairs compared to baseline. In contrast, PPO representations showed higher similarity for same-action states, but this separation emerged only in the later layers of the network hierarchy, with the strongest differentiation at the actor head. Notably, PPO representations did not show elevated similarity for mirrored state pairs.

\clearpage
\section{Further Results on Transfer Learning}

\begin{figure*}[tbh]
    \centering
    \includegraphics[width=\textwidth]{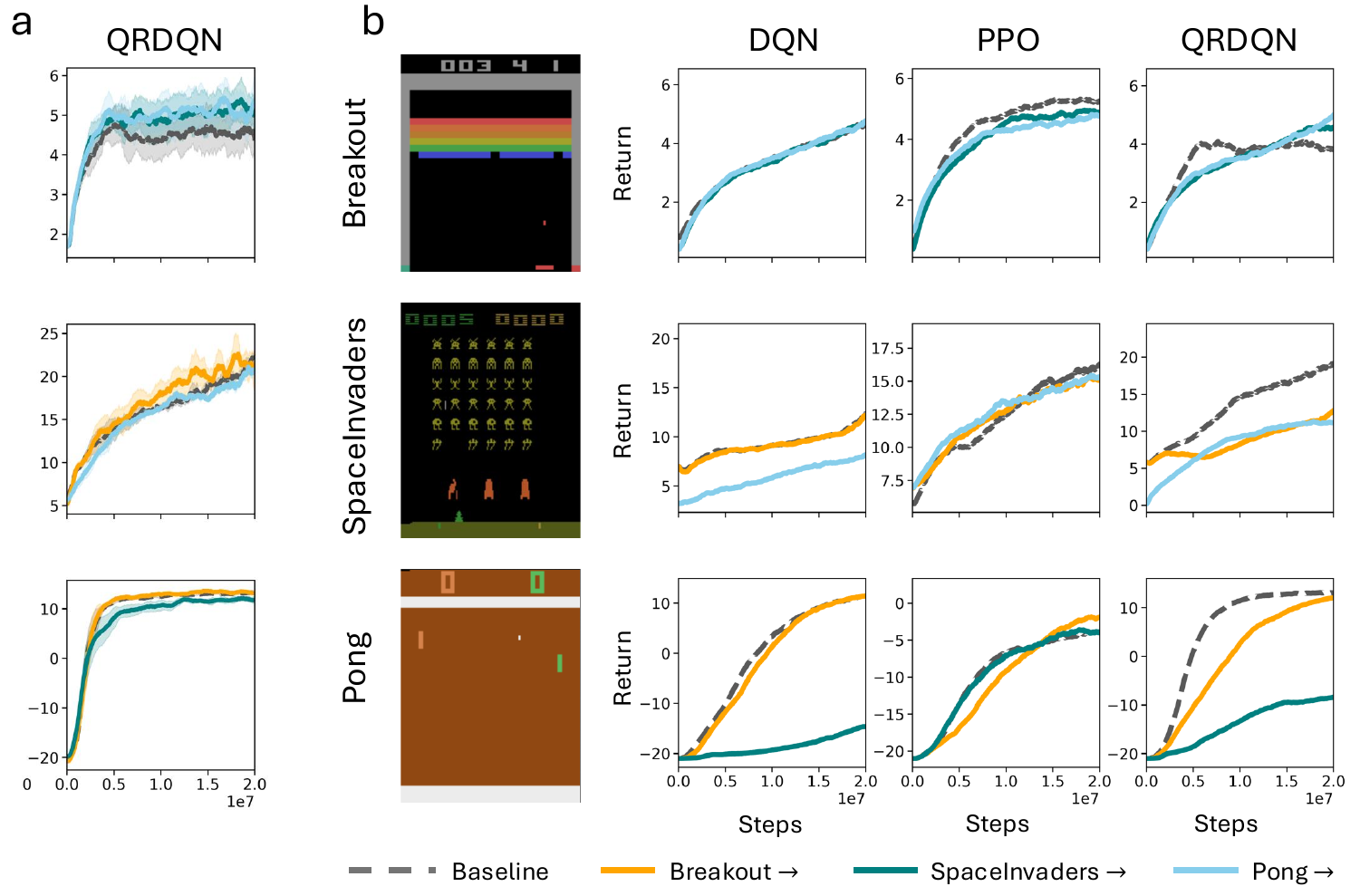}
    \caption{(a) These plots are as in~\autoref{fig:transfer_rebuttal_combined} in the main text but for the distributional RL algorithm QRDQN. It shows similar transfer performance to vanilla DQN. (b) Here we show an experiment similar to~\autoref{fig:transfer_rebuttal_combined} but rather than train on the target task after 10 million steps, we train on the source task for the full 200 million steps before transferring. In doing so we suspect many models enter the regime of plasticity loss~\cite{lee2022maslow, abbas2023loss, dohare2024loss} and thus show only neutral or negative transfer.
    }
    \label{fig:fulltransfer}
\end{figure*}

\begin{table}[!h]
\centering
\footnotesize
\caption{\textbf{Atari Transfer Results Table.} Here we take the final window of episode rewards for each experiment to understand the evidence for transfer effects. The table is computed by taking the final 200 episodes. We report the mean and standard deviations along with the differences from baseline to with pretraining (color coded). Visually we can see that DQN and to a lesser extent QRDQN benefit, while PPO does not as much.}
\label{tab:stats}
\begin{tabular}{lccccc}
\toprule
Model & Target & Source & Baseline & Pretrain & $\Delta\mu$ \\
\midrule
DQN  & Breakout & Pong & 4.15$\pm$0.8 & 4.90$\pm$1.4 & \cellcolor[HTML]{6FB56F} 0.75 \\
DQN  & Breakout & SpaceInvaders & 4.15$\pm$0.8 & 5.45$\pm$0.1 & \cellcolor[HTML]{006400}1.30 \\
DQN  & Pong & Breakout & 11.78$\pm$1.4 & 12.69$\pm$0.4 & \cellcolor[HTML]{4FA04F}0.91 \\
DQN  & Pong & SpaceInvaders & 11.78$\pm$1.4 & 5.89$\pm$1.9 & \cellcolor[HTML]{8B0000}-5.89 \\
DQN  & SpaceInvaders & Breakout & 19.27$\pm$2.1 & 19.29$\pm$3.0 & \cellcolor[HTML]{FDFEFD}0.01 \\
DQN  & SpaceInvaders & Pong & 19.27$\pm$2.1 & 20.17$\pm$2.3 & \cellcolor[HTML]{55A555}0.89 \\
PPO  & Breakout & Pong & 5.13$\pm$0.4 & 4.73$\pm$1.0 & \cellcolor[HTML]{F6EEEE}-0.41 \\
PPO  & Breakout & SpaceInvaders & 5.13$\pm$0.4 & 5.06$\pm$0.2 & \cellcolor[HTML]{FCFBFB}-0.07 \\
PPO  & Pong & Breakout & 2.12$\pm$2.9 & -1.67$\pm$0.6 & \cellcolor[HTML]{B24F4F}-3.98 \\
PPO & Pong & SpaceInvaders & 2.12$\pm$2.9 & -2.05$\pm$3.6 & \cellcolor[HTML]{AD4747}-4.17 \\
PPO & SpaceInvaders & Breakout & 15.72$\pm$0.3 & 15.17$\pm$4.0 & \cellcolor[HTML]{F2E6E6}-0.55 \\
PPO & SpaceInvaders & Pong & 15.72$\pm$0.3 & 16.82$\pm$1.5 & \cellcolor[HTML]{2E7F2E}1.10 \\
QRDQN & Breakout & Pong & 4.34$\pm$1.3 & 4.87$\pm$1.1 & \cellcolor[HTML]{9ED19E}0.54 \\
QRDQN & Breakout & SpaceInvaders & 4.34$\pm$1.3 & 4.73$\pm$1.0 & \cellcolor[HTML]{B7DEB7}0.39 \\
QRDQN & Pong & Breakout & 13.18$\pm$0.5 & 13.33$\pm$0.7 & \cellcolor[HTML]{E4F3E4}0.16 \\
QRDQN & Pong & SpaceInvaders & 13.18$\pm$0.5 & 11.79$\pm$1.2 & \cellcolor[HTML]{E2B8B8}-1.39 \\
QRDQN & SpaceInvaders & Breakout & 21.65$\pm$2.7 & 4.15$\pm$0.8 & \cellcolor[HTML]{5FAA5F}0.84 \\
QRDQN & SpaceInvaders & Pong & 21.65$\pm$2.7 & 20.15$\pm$2.4 & \cellcolor[HTML]{DEB1B1}-1.50 \\
\bottomrule
\end{tabular}
\end{table}

\clearpage
\section{Experimental Details}

\subsection{Labyrinth navigation}
\label{app:meister_descr}
We implemented the labyrinth task from \citet{rosenberg2021mice} as a gridworld with four discrete actions (up, down, left, right). Trying to walk into a wall resulted in no state change. The agent received a reward of +1 reward at the goal and 0 elsewhere. Episodes terminated after 200 steps or upon reaching the goal.

For both DQN and PPO, we used identical convolutional encoders: Conv (16 channels) → ReLU → Conv (32 channels) → ReLU → FC (128) → ReLU → FC (128). DQN output Q-values per action; PPO had separate policy and value heads with a shared encoder.

For training DQN, we used learning rate 3e-4, batch size 64, $\epsilon$-greedy exploration ($\epsilon$=1.0, decay 0.995), replay buffer 50K, target update every 50 steps, and optimistic initialization. For PPO, we used learning rate 3e-4, batch size 64, $\gamma$=0.99, $\lambda_{\text GAE}$=0.95, clip coefficient 0.2, entropy coefficient 0.01, 5 update epochs per rollout. DQN was trained for 10000 episodes, whereas PPO was trained for 4M steps. We used the AdamW optimizer for both.

For the representaitonal analysis, we extracted representations at each layer for all states, mean-centered, then computed pairwise cosine similarities. We derived MDP symmetry labels from homomorphism-induced equivalence classes and $\pi^*$ labels from optimal policy groupings.

\subsection{Cartpole}
\label{app:cartpole_descr}
We used the standard \texttt{CartPole-v1} from Gymnasium with 4-dimensional continuous state space and 2 discrete actions.
We used 2-layer MLPs with 256 hidden units per layer, ReLU activations. DQN output two Q-values; PPO completely separated actor and critic networks. Both implementations were based on stable-baselines configurations.

For the analysis, we sampled 500 random states and their mirror reflections (negating position and velocity). We computed cosine similarity between mirrored state pairs and compared them against random baseline pairs. We defined policy symmetry by states where the learned policy selects identical actions.

\subsection{Pong}
\label{app:pong_descr}
We used the \texttt{PongNoFrameskip-v4} environment from Gymnasium~\cite{brockman2016openai} with standard Atari preprocessing~\cite{mnih2013playing}. 
For the analysis, we used pre-trained DQN and PPO mdoels from the Stable Baselines3 model zoo \citep{stable-baselines}.
We sampled states during gameplay and manually constructed their vertical mirror images (exploiting top-bottom symmetry). We extracted representations from convolutional as well as fully-connected layers. PPO shared encoders between the actor and the critic, only differing at the respective heads.

\subsection{LLM Experiments}
\label{app:llm_descr}
We used the Qwen2.5-72B-Instruct model for our analysis.
We presented a smaller version of the abstract labyrinth MDP (cf.\ Figure~\ref{fig:meister}b) as a navigation graph and prompted the model to navigate step-by-step from each starting state to a fixed goal. 
We randomized the node numbering to prevent the model from extracting the graph structure directly from the node indices.
We used one overall shared prompt describing the general task followed by several different prompt formats describing the full graph structure in detail:

\begin{enumerate}
    \item \textbf{Directed Edge List (Ordered):} Each edge listed as \texttt{NodeA --Direction--> NodeB}, presented in hierarchical order following the tree structure from root to leaves. This format preserves structural information through ordering.
    
    \item \textbf{Directed Edge List (Randomized):} Identical edge notation to (1), but edges are randomly shuffled, removing structural cues from presentation order.
    
    \item \textbf{ASCII Tree Diagram:} A visual text representation using indentation and ASCII characters to show the hierarchical tree structure. Node relationships and directions are explicitly displayed in a spatially organized format.
    
    \item \textbf{Adjacency List (JSON):} A dictionary mapping each node to its neighbors with direction labels. Both node order and direction order within each entry are randomized.
    
    \item \textbf{Relation Triples:} Explicit directional predicates of the form \texttt{Direction(from, to)}, e.g., \texttt{North(7, 3)}. Triples are listed in random order.
    
    \item \textbf{Path Mapping:} Each node described by its path from the start node using cardinal directions, e.g., \texttt{Node 11: East-South-East-South}. Mappings listed in random order.
\end{enumerate}

The different prompt formats are illustrated in \autoref{fig:prompt_formats}. For analysis, we recorded hidden states from all transformer layers after the final token specifying the start state. The analysis followed the same RSA procedure as for RL agents.
In \autoref{fig:llm_sim}, we show the representation similarities only for the ASCII Tree Diagram (top) and Directed Edge List (Randomized) (bottom) formats, whereas
\autoref{fig:llm_sim_full} shows all six formats tested.

\begin{figure*}[t]
\centering
\begin{minipage}[t]{0.32\textwidth}
\begin{lstlisting}[basicstyle=\tiny\ttfamily,frame=single,mathescape=true,title={\small (a) Edge List (Ordered)}]
9 --North--> 14
9 --South--> 8
14 --South--> 9
8 --North--> 9
14 --West--> 7
14 --East--> 15
7 --East--> 14
15 --West--> 14
8 --West--> 13
8 --East--> 6
...

You are currently at Node $x$.
Your goal is to reach Node 11.
\end{lstlisting}
\end{minipage}
\hfill
\begin{minipage}[t]{0.32\textwidth}
\begin{lstlisting}[basicstyle=\tiny\ttfamily,frame=single,mathescape=true,title={\small (b) Edge List (Randomized)}]
3 --South--> 7
8 --West--> 13
8 --North--> 9
6 --West--> 8
13 --East--> 8
12 --South--> 13
11 --North--> 6
7 --North--> 3
14 --East--> 15
4 --South--> 15
...

You are currently at Node $x$.
Your goal is to reach Node 11.
\end{lstlisting}
\end{minipage}
\hfill
\begin{minipage}[t]{0.32\textwidth}
\begin{lstlisting}[basicstyle=\tiny\ttfamily,frame=single,mathescape=true,title={\small (c) ASCII Tree Diagram}]
[9] (Start)
  |
  +--North-->
    [14] (came from North)
      |
      +--West-->
        [7] (came from West)
          |
          +--North-->
            [3] (came from North)
          |
          +--South-->
            [10] (came from South)
      ...

You are currently at Node $x$.
Your goal is to reach Node 11.
\end{lstlisting}
\end{minipage}

\vspace{0.5em}

\begin{minipage}[t]{0.32\textwidth}
\begin{lstlisting}[basicstyle=\tiny\ttfamily,frame=single,mathescape=true,title={\small (d) Adjacency List (JSON)}]
{
  "1": {
    "North": 13
  },
  "10": {
    "North": 7
  },
  "6": {
    "North": 2,
    "West": 8,
    "South": 11
  },
  "8": {
    "West": 13,
    "North": 9,
    "East": 6
  },
  ...
}

You are currently at Node $x$.
Your goal is to reach Node 11.
\end{lstlisting}
\end{minipage}
\hfill
\begin{minipage}[t]{0.32\textwidth}
\begin{lstlisting}[basicstyle=\tiny\ttfamily,frame=single,mathescape=true,title={\small (e) Relation Triples}]
North(7, 3)
South(12, 13)
South(7, 10)
East(8, 6)
North(5, 15)
South(3, 7)
South(9, 8)
North(9, 14)
North(10, 7)
North(8, 9)
...

You are currently at Node $x$.
Your goal is to reach Node 11.
\end{lstlisting}
\end{minipage}
\hfill
\begin{minipage}[t]{0.32\textwidth}
\begin{lstlisting}[basicstyle=\tiny\ttfamily,frame=single,mathescape=true,title={\small (f) Path Mapping}]
Node 9: Start
Node 5: East-North-East-South
Node 1: East-South-West-South
Node 15: East-North-East
Node 10: East-North-West-South
Node 8: East-South
Node 14: East-North
Node 2: East-South-East-North
Node 6: East-South-East
Node 3: East-North-West-North
...

You are currently at Node $x$.
Your goal is to reach Node 11.
\end{lstlisting}
\end{minipage}

\caption{\textbf{Six graph representation formats for LLM navigation experiments.} All formats encode the same underlying tree structure (15 nodes, depth 4) with randomized node IDs and start node $x$ is varied to generate state representations. (a)~Directed edges in hierarchical order preserving tree structure. (b)~Same edges as (a) but randomly shuffled. (c)~ASCII visualization with explicit spatial hierarchy. (d)~JSON adjacency list with randomized node/direction order. (e)~Predicate-based directional relations in random order. (f)~Nodes identified by cardinal direction paths from start.}
\label{fig:prompt_formats}
\end{figure*}

\begin{figure*}[tbh]
    \centering
    \includegraphics{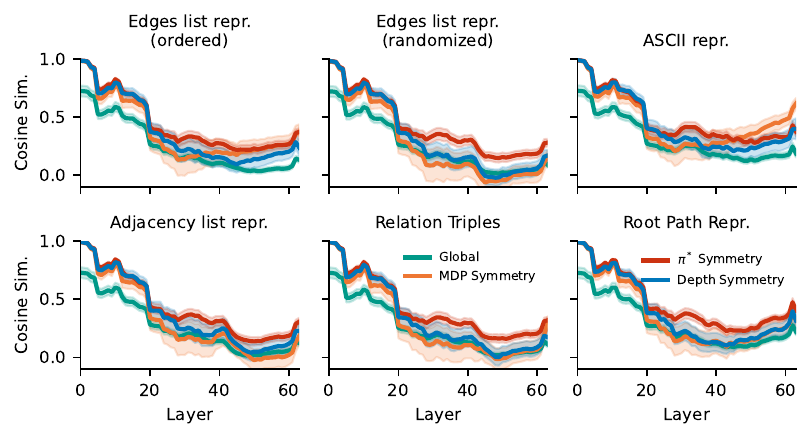}
    \caption{\textbf{Representational symmetry structure in an LLM for different prompt formats.}
    Same as \autoref{fig:llm_sim} but for all the prompt formats in \autoref{fig:prompt_formats}.
    Policy symmetry is consistently visible in the representations across all formats, but elevated MDP symmetry is only observed with prompts that already contain some structural information like the ASCII representation or, to a lesser extent, in the ordering of the graph edges in the ordered Edge List.
    }
    \label{fig:llm_sim_full}
\end{figure*}

\subsection{Atari Transfer}
\label{app:transfer_descr}
We used open source implementations from~\citet{stable-baselines} for PPO, DQN and QRDQN with their default architectures and hyperparameter configurations. Further, we used the standard Atari pre-processing via stable-baselines wrappers. For the results shown in~\autoref{fig:transfer_rebuttal_combined}, we trained a given model on the source task for 10 million steps before initialising a new head and training on the target task. Each run reported there is for a total of 20 million steps.

\subsection{Computational Resources}
\label{app:transfer_compute}
Atari Deep RL experiments are resource intensive. We trained 3 different models (PPO, DQN, QRDQN) on 3 source tasks, and then a further two target tasks for each source run. With 3 random seeds this amounts to a total 81 experiments. Experiment were run on a cluster with H100 and A100 GPUs for a sum total of approximately 1600 GPU hours. Further experiments shown in~\autoref{fig:fulltransfer} were only a single seed and did not require separate source runs since the baseline checkpoints were taken from stable-baselines, but still contributed in addition to this estimate. As the statistics in~\autoref{tab:stats} show, more seeds would definitely be preferable, although 3 seeds are commonplace in deep RL papers across major ML conferences.

\section{Extended Discussion of LLM Results}
\label{app:llm_discussion}

The representational structure learned by the LLM is highly sensitive to how the task is described. When the task is presented using an ASCII representation, the LLM shows elevated similarity for states equivalent under the MDP homomorphism, comparable to what we observed in DQN. This suggests the model is able to extract and represent the abstract task structure when it is made explicit in the prompt format. However, when the same task is described as an edge list, this MDP symmetry largely disappears, with similarity levels closer to the naive depth-based grouping, wherein states at the same level of the tree are grouped together.

Across both prompt formats, the LLM consistently shows high similarity for states sharing the same optimal action, indicating that action-level information is robustly encoded regardless of task description. This high policy similarity emerges early in the network and persists through all layers, similar to the pattern we observed in PPO.

These findings demonstrate that the symmetries encoded in LLM representations during in-context task execution are not fixed properties of the model, but depend critically on how the task is presented. This prompt-dependent encoding raises an interesting question for LLM post-training via RLHF: RL post-training using policy-gradient methods (PPO, GRPO) may shift LLM representations towards policy-like action symmetries, potentially at the cost of the structural invariances that facilitate in-context generalization---an effect analogous to what we observe for PPO in Atari transfer.
A systematic investigation of how RLHF post-training shapes representational invariances in LLMs is an important direction for future work.


\newpage

\end{document}